\documentclass[letterpaper]{article}

\usepackage{arxiv}

\usepackage[utf8]{inputenc}
\usepackage[T1]{fontenc}

\usepackage{amsmath,amssymb,amsthm,mathtools,bm}
\usepackage{empheq}
\usepackage{siunitx}
\usepackage{graphicx}
\usepackage{booktabs}
\usepackage{caption}
\usepackage{subcaption}
\usepackage{enumitem}
\usepackage{tikz}
\usetikzlibrary{calc,arrows.meta,angles,quotes,positioning}

\usepackage{url}
\usepackage{microtype}
\usepackage{doi}

\usepackage{algorithm}
\usepackage{algpseudocode}
\usepackage[numbers,sort&compress]{natbib}

\theoremstyle{definition}

\theoremstyle{plain}

\definecolor{teal}{HTML}{0F6E56}
\definecolor{coral}{HTML}{C0451F}
\definecolor{bluec}{HTML}{185FA5}
\definecolor{amberc}{HTML}{BA7517}
\definecolor{purplec}{HTML}{534AB7}
\definecolor{myorange}{HTML}{E8743B}
\definecolor{mygreen}{HTML}{2E9E4F}
\definecolor{mypurple}{HTML}{9B36B8}
\definecolor{mylightblue}{HTML}{9AD0F0}
\definecolor{myblue}{HTML}{1F6FB2}

\title{Real-Time sEMG-Based Telecontrol of an Assistive Robotic Arm Using a 1D Convolutional Neural Network}

\author{
Edgar Manacorda \\
Department of Mechanical Engineering \\
Polytechnique Montréal \\
Montréal, Québec, Canada \\
\texttt{edgar.manacorda@etud.polymtl.ca}
\And
Mena Samir Kama Abouseffien \\
Department of Mechanical Engineering \\
Polytechnique Montréal \\
Montréal, Québec, Canada \\
\texttt{mena-samir-kamal.abouseffien@etud.polymtl.ca}
\And
Olivier Lecompte \\
Department of Mechanical Engineering \\
Polytechnique Montréal \\
Montréal, Québec, Canada \\
\texttt{olivier.lecompte@etud.polymtl.ca}
\And
Amandine Gesta \\
Department of Mechanical Engineering \\
Polytechnique Montréal \\
Montréal, Québec, Canada \\
\texttt{amandine.gesta@etud.polymtl.ca}
\And
\href{https://orcid.org/0000-0003-2101-9651}
{\includegraphics[scale=0.06]{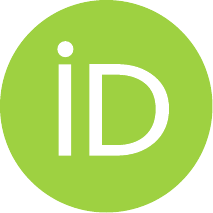}\hspace{1mm}Abolfazl Mohebbi}$^{*}$ \\
Department of Mechanical Engineering \\
Polytechnique Montréal \\
Montréal, Québec, Canada \\
\texttt{abolfazl.mohebbi@polymtl.ca}
}

\renewcommand{\shorttitle}{Real-Time sEMG-Based Telecontrol of an Assistive Robotic Arm}

\hypersetup{
colorlinks=true,
linkcolor=blue!50!black,
citecolor=blue!50!black,
urlcolor=blue!50!black,
pdftitle={Real-Time sEMG-Based Telecontrol of an Assistive Robotic Arm Using a 1D Convolutional Neural Network},
pdfsubject={surface electromyography, robotic teleoperation, deep learning, human-machine interface},
pdfauthor={Edgar Manacorda, Abolfazl Mohebbi},
pdfkeywords={surface electromyography, gesture recognition, convolutional neural network, robotic teleoperation, real-time control},
}

\begin{document}

\maketitle

\renewcommand{\thefootnote}{*}
\footnotetext{Corresponding author. Associate Professor, Director of the Polytechnique Lab for Assistive and Rehabilitation technologies (POLAR), Polytechnique Montréal. Email: \texttt{polar@polymtl.ca}.}
\renewcommand{\thefootnote}{\arabic{footnote}}

\begin{abstract}
Motor impairments affecting the upper limb significantly reduce autonomy in daily activities, particularly for tasks involving object manipulation. Assistive robotic arms offer a promising solution, provided they can be controlled in an intuitive, reliable, and responsive manner. Among human--machine interface approaches, surface electromyography (sEMG) enables non-invasive access to muscle activity and thus to the user's motor intentions. This work proposes a real-time sEMG-based interface for the teleoperation of an assistive robotic arm. The system relies on four-channel sEMG acquisition, signal preprocessing, segmentation into sliding windows, and classification using a one-dimensional convolutional neural network (CNN). Several real-time strategies are investigated, including threshold-based onset detection, a two-stage classification approach (rest vs movement followed by gesture recognition), and a single classifier handling both rest and five gestures. The complete pipeline is implemented and evaluated both in simulation and on a real robotic platform. The CNN-based approach achieves high classification performance, with a test accuracy above 90\% and strong generalization on experimentally acquired signals. The system exhibits stable real-time behavior, with an average latency of approximately \SI{0.32}{\second} consistent with the chosen windowing strategy, and the robot can be controlled reliably using discrete gestures, producing coherent and smooth movements in both simulated and real environments. These findings demonstrate the feasibility of sEMG-based telecontrol for assistive robotics and highlight the importance of integrating signal processing, deep learning, and control strategies within a unified real-time framework. Future work may explore hybrid control approaches combining sEMG with additional sensing modalities to further improve robustness and usability.
\end{abstract}


\section{Introduction}
\label{sec:intro}

Upper limb motor impairments, whether of neurological, traumatic, or neuromuscular origin, directly affect autonomy in activities of daily living. Simple actions such as reaching for an object, grasping it, repositioning it, or handling a utensil become difficult, slow, or impossible to perform without external assistance. In this context, assistive robotic devices offer a promising solution: they can partially compensate for the loss of upper limb function and reduce dependence on a caregiver. However, the clinical and functional value of a robotic arm depends not only on its mechanical capabilities but also on the quality and intuitiveness of the interface used to control it~\citep{chung2013functional, kim2012autonomy}.

Current solutions for controlling assistive devices rely on a variety of strategies, including joysticks, position-sensor controls, visual interfaces, eye tracking, shared control, and physiological interfaces. Each has advantages and significant limitations. Conventional joystick interfaces can be effective when the user retains sufficient residual motor function, but they remain relatively unnatural and are not always suitable for individuals with severe impairments. Visual or environment-based approaches may improve certain tasks but depend on perception conditions that are difficult to guarantee in practice. These limitations explain the growing interest in physiological interfaces capable of translating movement intention more directly into robotic commands~\citep{chung2013functional, bi2019review, sarhan2023review}. Among these approaches, surface electromyography (sEMG) occupies an important place. It non-invasively measures the electrical activity produced by muscles during a motor intention or execution, which makes it a relevant candidate for human--machine interfaces, especially for detecting simple gestures or movement intentions at the level of the hand, wrist, or forearm. Recent literature shows that sEMG is used across a variety of fields, ranging from myoelectric prostheses to rehabilitation devices and robotic teleoperation systems. Its main advantage is that it provides a signal directly related to muscle activation, with relatively low-cost acquisition compatible with real-time use~\citep{alcan2023current, bi2019review, yadav2023recent}.

Despite this interest, controlling robots with sEMG remains difficult. The signal is sensitive to noise, electrode displacement, muscle fatigue, inter-individual and inter-session variability, and general acquisition conditions, so signal quality and stability over time are never guaranteed. This variability is one of the main barriers to robust use of sEMG outside strictly controlled conditions. In gesture recognition specifically, electrode positioning directly influences performance, which highlights the importance of consistency between model training and real-world operating conditions~\citep{bi2019review, yadav2023recent, wang2023effects}. At the same time, gesture recognition from sEMG has evolved rapidly through machine learning and, more recently, deep learning. Older approaches relied mainly on manual feature extraction followed by a conventional classifier, whereas deep architectures, particularly convolutional neural networks, now occupy a central place in the literature. Several studies have shown that a one-dimensional (1D) CNN can effectively exploit the temporal and multichannel structure of sEMG signals for hand gesture recognition, and others have explored hybrid architectures combining convolutions and sequential models, such as CNN--RNN models, to better account for temporal dynamics. These results show that deep learning can achieve high classification performance, but they often concern offline evaluations on well-structured datasets~\citep{atzori2016deep, hu2018novel, li2021gesture}.

However, good offline classification performance alone does not guarantee satisfactory real-time robotic control. Moving from a classification model to a complete telecontrol system introduces additional constraints: the signals must be processed causally, the transition from rest to movement must be robustly detected, the data must be segmented into inference-compatible windows, decisions must be produced at a sufficiently high frequency, and those decisions must be transformed into stable robotic commands. To this is added system latency; a system that is too slow, even if accurate, may feel unnatural or difficult to use~\citep{bi2019review, sarhan2023review, yadav2023recent}.

Furthermore, work on upper-limb robots shows that assistive and rehabilitation robotics are closely related but distinct fields. Much of the literature focuses on post-stroke rehabilitation or therapeutic robots, with encouraging but sometimes mixed results depending on the clinical context, the tasks evaluated, and the selected metrics~\citep{moulaei2023overview, nice2023robot, huo2023effects}. This confirms that mechanical performance alone is not sufficient: the quality of the human--robot interaction and the relevance of the control interface remain critical. From the perspective of an assistive robot, the challenge is to provide control that is simple, reliable, and intuitive enough to be usable in functional tasks, even if the system initially controls only a limited number of degrees of freedom~\citep{chung2013functional, kim2012autonomy}.

It is within this context that the present work is situated. The objective is to develop a human--machine interface based on sEMG signals to enable telecontrol of a 7 degrees-of-freedom (DOF) robotic arm through hand and wrist movements. More specifically, the project aims to transform muscle activity measured on the forearm into simple Cartesian commands applied to a robotic arm, first in simulation and then on a real robot. The work includes the implementation of a complete real-time pipeline covering acquisition, preprocessing, muscle activation detection, window segmentation, normalization, convolutional neural network inference, and robotic control, making it possible to evaluate not only gesture recognition but also its effective integration into a robotic control loop. The main hypothesis of this work is that a system based on multichannel sEMG signals and a classifier can enable reliable and sufficiently responsive control of an assistive robotic arm through several discrete commands. More specifically, it is assumed that by combining coherent preprocessing, appropriate segmentation, a robust muscle activation detection strategy, and a stabilized decision logic, it is possible to obtain interpretable and usable robotic behavior under semi-real-time conditions. Based on this hypothesis, the present work compares several processing strategies, selects a relevant architecture, and validates the complete pipeline in a robotic telecontrol context.
\section{Methods}
\label{sec:methods}

\subsection{Materials}
\label{subsec:materials}

The sEMG signals are acquired using a network of four Delsys wired electrodes and a reference electrode connected to a Delsys Bagnoli-4 EMG amplifier with a bandwidth of \SIrange{20}{2000}{\hertz}~\citep{delsys2003bagnoli}. The gain is set to 1000 for each of the four channels. Each channel is connected via a coaxial cable to the corresponding input on a National Instruments cDAQ-9171 CompactDAQ chassis, a compact and portable data-acquisition system~\citep{ni_cdaq9171}, which is connected by USB to a laptop where the data are acquired at a sampling rate of \SI{2000}{\hertz}. The robot is the UFACTORY xArm7, a 7-DoF robotic arm~\citep{ufactory_xarm}. The numerical simulation is a replica of the xArm7 created using Robotics Toolbox for Python~\citep{corke2021robotics}.

\subsection{Gesture Classification}
\label{subsec:classification}

The objective of this stage is to develop a model capable of classifying hand gestures from multichannel sEMG signals. In addition to the \emph{rest} state, five gestures were considered: extension, flexion, ulnar deviation, radial deviation, and grip.

\subsubsection{Database}
\label{subsubsec:database}

\begin{figure}[htbp]
  \centering
  \includegraphics[width=0.7\linewidth]{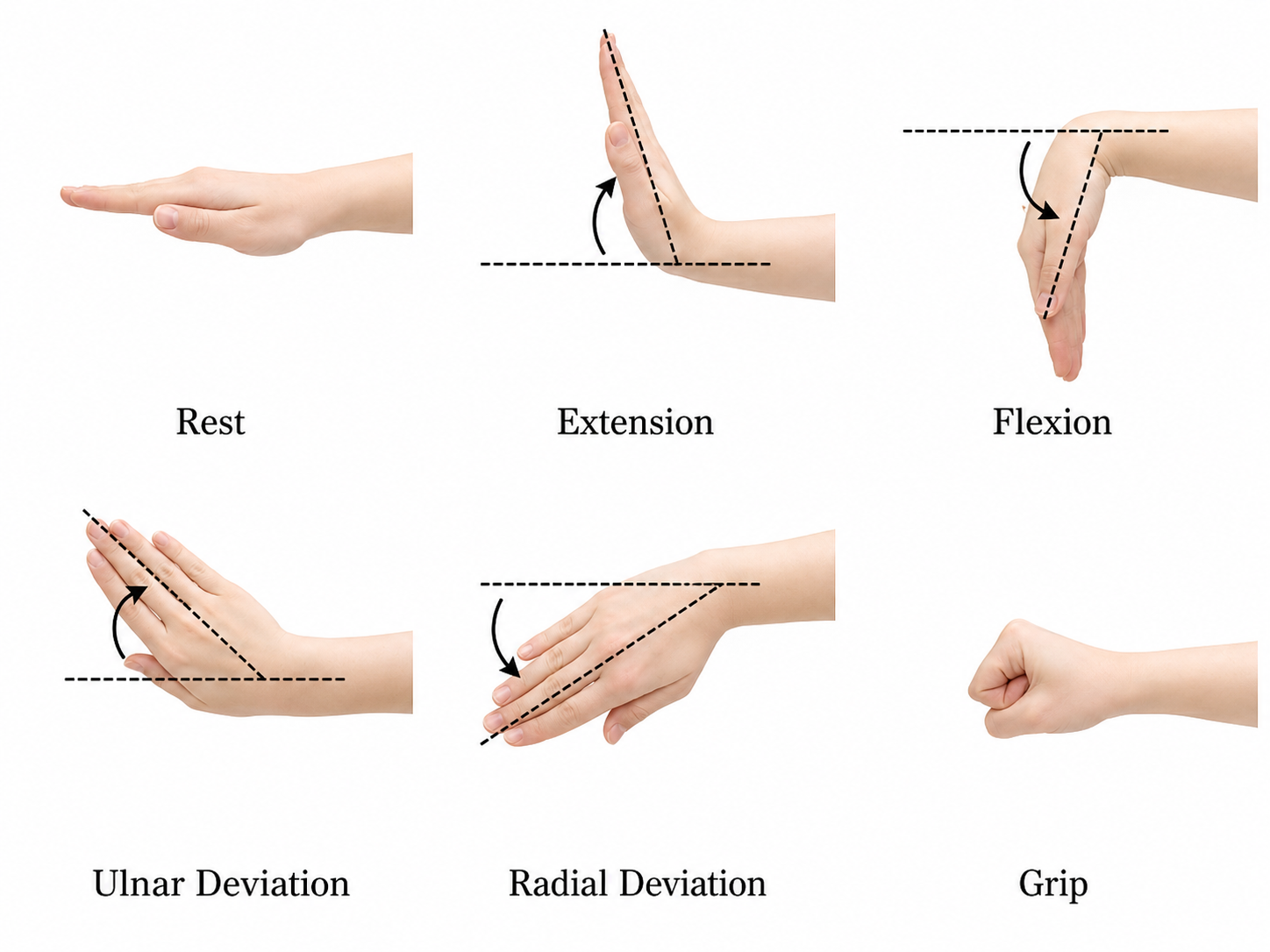}
  \caption{Gestures of interest used for sEMG classification. Reproduced from~\citep{ozdemir2022dataset}.}
  \label{fig:gestures}
\end{figure}

All data used in this work were obtained from a publicly available online database in order to access a large quantity of data. The database, created by Ozdemir et al., includes four-channel sEMG recordings from 40 participants with an equal distribution of genders~\citep{ozdemir2022dataset}. Each participant performed 10 gestures with 5 repetitions per gesture. Among these, six were selected for this work: rest, extension, flexion, ulnar deviation, radial deviation, and grip (Figure~\ref{fig:gestures}). Each gesture is therefore represented by 200 recordings, for a total of 1{,}200 CSV files. Each recording corresponds to a single six-second trial sampled at \SI{2000}{\hertz} across four EMG channels. The signals had already been filtered with a \SI{50}{\hertz} notch filter to remove power-line interference, followed by a 6th-order Butterworth band-pass filter between \SI{5}{} and \SI{500}{\hertz}. % No normalization or MVC calibration had been applied.

\subsubsection{Onset Detection}
\label{subsubsec:onset}

The 1{,}200 files are assumed to contain only the signals corresponding to the associated gestures. However, some participants took time before starting their movements, and the automatic segmentation proposed by the dataset authors does not account for this human delay. The automatically segmented files are therefore noisy, as rest may have been included in the first milliseconds of each gesture. Providing these files directly to the classification models would label rest as a gesture and introduce noise into training. To isolate meaningful segments while avoiding manual sorting or arbitrary cutting, a muscle-activity onset-detection algorithm based on an adaptive threshold applied to the signal envelope was developed and applied to each recording.

For each file, the algorithm first computes the envelope of each of the four channels using a moving RMS with a smoothing window of \SI{25}{\milli\second}. It then selects the most active channel---the one measuring the highest muscle activity for the given gesture---based on the RMS envelope values. This channel is used for detection, and once activation is detected its position is replicated across the other channels of the same file.

Two thresholding methods were compared on the RMS envelope of the selected channel. The first defines an adaptive threshold from the distribution of local maxima (activity peaks), using percentiles ranging from 70\% to 90\% depending on the tests. The second defines a threshold from a baseline considered to be at rest. After several tests, the second method proved better, as the local-maxima percentile method depends too strongly on signal content and is too sensitive to artifacts; it was therefore selected.

For this method, a robust baseline was estimated from the median and the median absolute deviation (MAD) of the first \SI{500}{\milli\second}, considered to be rest. The threshold was defined as
\begin{equation}
  \mathrm{threshold} = \mathrm{median} + k\cdot\sigma,
  \label{eq:onset}
\end{equation}
where $k = 2.5$ was chosen empirically and $\sigma$ is the robust estimate of the standard deviation derived from the MAD. An onset was then searched for in the region from \SI{100}{} to \SI{1000}{\milli\second}, a temporal constraint imposed to restrict the search window and avoid false detections, and determined empirically to give the best results. An onset was considered detected when the signal exceeded the threshold for a minimum duration of \SI{20}{\milli\second}; a \SI{50}{\milli\second} margin was then added before the onset to preserve part of the transition phase, and all preceding signal was removed from the four channels as rest.

When no onset was detected, a fallback strategy was applied, either by selecting a fixed cutoff point or by keeping the full signal, depending on the configuration. Analysis of the undetected files suggests that the main cause is a baseline too close to the activity level, indicating that the participant likely began the movement before acquisition started, so that the file contains only the gesture. Classification performance was better when the fixed-cutoff fallback was used: although it removes a few milliseconds that belong to the gesture, it likely removes unobserved noise, ultimately improving performance.

\subsubsection{Dataset Construction}
\label{subsubsec:dataset}

After temporal cropping based on onset detection, the remaining signal of each file is segmented into sliding windows, each representing an independent training instance. This both increases the number of examples and brings the learning task closer to the intended real-time operation, in which decisions are made locally on successive windows. Several configurations were tested to balance the information contained in each window, prediction stability and accuracy, and potential latency:
\begin{itemize}[nosep]
  \item a \SI{250}{\milli\second} window (500 samples at \SI{2000}{\hertz}) with 50\% overlap;
  \item a \SI{300}{\milli\second} window (600 samples) with a \SI{75}{\milli\second} step (75\% overlap);
  \item a \SI{350}{\milli\second} window (700 samples) with 75\% overlap.
\end{itemize}
The second configuration was retained: each window contains 600 samples, and successive windows are spaced by 150 samples. Shorter windows reduced observable muscular information and made classification less stable, whereas longer windows increased latency and reduced responsiveness without improving accuracy. Incomplete windows at the end of a signal were discarded to ensure a fixed input size. For a file representing one repetition of a gesture, with an average duration of \SI{5.5}{\second} after onset detection, the number of windows per file is
\begin{equation}
  N_{\text{windows}} = 1 + \frac{5.5\,\text{s}\times 2000\,\text{Hz} - 0.3\,\text{s}\times 2000\,\text{Hz}}{0.075\,\text{s}\times 2000\,\text{Hz}} \approx 70,
  \label{eq:nwindows}
\end{equation}
so the total number of windows generated per gesture is approximately
\begin{equation}
  N_{\text{total}} = 70\ \text{windows} \times 40\ \text{participants} \times 5\ \text{cycles/gesture} = 14{,}000\ \text{windows per gesture}.
  \label{eq:ntotal}
\end{equation}

The original dataset contains six classes: \texttt{g0} rest, \texttt{g1} extension, \texttt{g2} flexion, \texttt{g3} ulnar deviation, \texttt{g4} radial deviation, and \texttt{g5} grip. Several variants of the classification problem were considered to evaluate different control strategies: a 6-class classifier including rest, a 5-class classifier of active gestures only, a 4-class classifier (extension, flexion, ulnar, radial), and a 2-class classifier for targeted comparisons. Finally, a binary (rest vs move) classifier was defined to later serve as an activity detector within a two-stage strategy. All windows extracted from a given file inherit the label of the source file, based on the assumption that, after onset-based segmentation, the remaining segment predominantly corresponds to the intended gesture. Although imperfect when onset detection is suboptimal, this assumption proved sufficiently robust to build a consistent dataset.

Because sEMG signals exhibit high inter-subject variability, the split was performed at the subject level rather than randomly at the window or file level; a random split would cause information leakage by placing data from the same individual in both training and test sets. Subjects 1--35 formed the training set and subjects 36--40 the test set, allowing evaluation on entirely unseen subjects. Within the training set, 20\% of the windows were reserved for validation using class-wise stratification, in order to monitor training, select the best model, and reduce overfitting.

\subsubsection{Classification Models}
\label{subsubsec:models}

Two families of approaches are generally considered for sEMG gesture recognition: classical machine learning with explicit feature extraction, and deep learning that learns representations directly from the raw or filtered signal. Recent literature shows growing interest in deep learning, particularly convolutional architectures, which often outperform handcrafted-feature approaches when signal variability is significant~\citep{li2021gesture, kok2024machine, gopal2022systematic}. Experimental efforts were therefore focused on deep architectures rather than on a parallel pipeline of manual feature extraction with classical models such as SVM, KNN, or decision trees. Two families were explored: a 1D CNN and an LSTM.

The first approach, ultimately retained, is a 1D CNN taking filtered signal windows as input. This preserves the raw nature of the signal while benefiting from prior filtering and onset-based segmentation, avoiding premature summarization and allowing the network to learn temporal patterns directly. A second variant replaced the filtered signal with its RMS envelope to provide a smoother input; however, performance was lower than with the raw filtered signal, suggesting that fine temporal details carry useful discriminative information partially lost in the envelope. A third family used LSTM networks to capture temporal dependencies between successive windows: \SI{300}{\milli\second} windows with 75\% overlap were grouped into sequences of five consecutive windows. With a \SI{75}{\milli\second} step, a sequence of five windows covers
\begin{equation}
  300 + (5-1)\times 75 = 600\ \si{\milli\second}.
  \label{eq:lstmseq}
\end{equation}
This provided richer temporal context but introduced a higher minimum latency and, in the experiments, performed worse than the 1D CNN with either input. It was therefore not retained. Before training, the data were normalized using per-channel z-score normalization. The mean and standard deviation were computed exclusively from the training-set windows, independently for each channel, and applied to the validation set, test set, and later to offline and real-time acquired data. Restricting the statistics to the training set avoids information leakage and keeps the measured performance representative of a realistic scenario, while ensuring consistency between training and deployment.

The final model is a lightweight yet sufficiently deep 1D CNN with three successive convolutional blocks: (i) a 1D convolution from 4 to 32 channels, kernel size 7, followed by batch normalization, ReLU, and max pooling; (ii) a convolution from 32 to 64 channels, kernel size 5, followed by batch normalization, ReLU, and max pooling; and (iii) a convolution from 64 to 128 channels, kernel size 5, followed by batch normalization, ReLU, and max pooling. A global adaptive average-pooling layer then reduces the temporal dimension to a compact representation, which is fed to a classification head composed of a linear layer from 128 to 64, a ReLU, a dropout layer with rate 0.5, and a final linear layer projecting to the number of classes. 1D convolutions suit multichannel time-series analysis and extract local patterns related to muscle dynamics; batch normalization stabilizes training; max pooling reduces the temporal dimension while adding robustness to small local variations; and global average pooling with dropout limits parameters and overfitting. The CNN was trained in PyTorch with the cross-entropy loss and the Adam optimizer, using a learning rate of \num{e-3}, a weight decay of \num{e-4}, a batch size of 128, and typically 30 epochs. At each epoch, performance was monitored on the validation set, and the model with the highest validation accuracy was saved for final evaluation, retaining the best trade-off observed during training rather than the final epoch. Increasing the number of epochs or adding early stopping did not improve final performance; the best results were consistently obtained with simpler configurations and a relatively short 30-epoch training.

The configurations were compared using validation accuracy, test accuracy, classification report (precision, recall, F1-score), confusion matrix, and, more qualitatively, compatibility with a future real-time pipeline. Selection was therefore based on the overall coherence of the trade-off between performance, simplicity, and usability rather than raw accuracy alone. The 1D CNN applied directly to the filtered signal, with a \SI{300}{\milli\second} window and 75\% overlap, emerged as the best solution, outperforming the RMS-envelope variants and the LSTM approaches while remaining simple enough for real-time integration. The dataset construction, windowing, normalization, and model selection proved interdependent: the best results came not from a highly complex model but from a coherent pipeline of reliable onset-based segmentation, appropriate windowing and normalization, and a well-designed 1D CNN architecture.

\subsection{Complete Real-Time System Pipeline}
\label{subsec:pipeline}

\begin{figure}[htbp]
  \centering
  \includegraphics[width=0.8\linewidth]{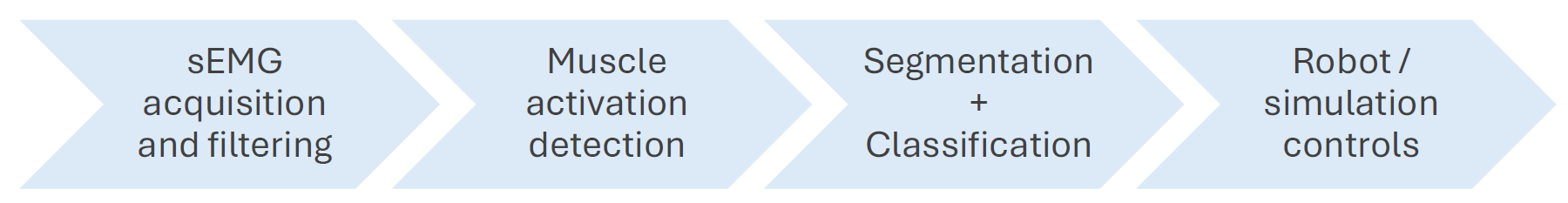}
  \caption{Real-time system pipeline: sEMG acquisition and filtering, muscle activation detection, segmentation and classification, and robot/simulation control.}
  \label{fig:pipeline}
\end{figure}

The real-time system aims to continuously transform, with minimal latency, the muscle activity measured at the forearm surface into control commands for the robotic arm. This requires sequential processing of the data as they are acquired. The pipeline receives raw sEMG signals, preprocesses them, detects muscle activity, segments the signal into windows, performs inference with the trained CNN, and transmits the predictions to a robotic control module (Figure~\ref{fig:pipeline}).

\subsubsection{sEMG Signal Acquisition}
\label{subsubsec:acquisition}

Surface EMG signals are continuously acquired on four channels using the NI-DAQ system at a sampling frequency of \SI{2000}{\hertz}, consistent with the source dataset. The electrode placement used for the real-time experiments was kept identical to that of the study used to build the training dataset, to maximize consistency between training data and online-control data, since sEMG performance strongly depends on the correspondence between training and usage conditions. Acquisition thus provides a continuous stream of raw, time-synchronized sEMG data for the subsequent processing steps.

\subsubsection{Real-Time Signal Preprocessing}
\label{subsubsec:preprocessing}

Preprocessing in real time follows the same principles as those used for dataset preparation and model training, to maintain methodological consistency between training and deployment. It first applies a \SI{60}{\hertz} notch filter to attenuate power-line interference, followed by a 6th-order Butterworth band-pass filter between \SI{5}{} and \SI{500}{\hertz}. This band corresponds to the standard range of interest for sEMG, removing low-frequency drifts and a portion of high-frequency noise that does not carry useful information for gesture recognition~\citep{muceli2024tutorial, boyer2023reducing}. The implementation relies on streaming filtering, applied progressively to each new block of acquired data. In the first version of the pipeline, an RMS envelope is also computed from the filtered signal as a scalar measure of muscle activity; it is not used as a direct input to the CNN, but only for muscle activation detection in the threshold-based strategy. The signal fed to the classifier remains the raw filtered signal, consistent with training. Preprocessing is therefore critical not only for signal quality but also for ensuring that the data presented to the model in real time exhibit statistical characteristics consistent with those seen during training.

\subsubsection{Muscle Activation Detection Strategies}
\label{subsubsec:strategies}

To convert muscle activity into usable commands, three strategies were developed and compared, differing mainly in how they identify the transition between rest and movement.

\emph{(a) Threshold-based detection.} The transition from rest to movement is determined using a threshold applied to the RMS envelope, following the principle of Section~\ref{subsubsec:onset}. The main difference is that the resting baseline is estimated from an initial five-second calibration phase during which the user remains at rest. The system starts in a rest state; as long as the activity remains below the threshold, no classification is performed. When the signal exceeds the threshold consistently over a sufficient number of consecutive samples, the system transitions to the \textsc{move} state and windows are processed by the classifier.

\emph{(b) Two-stage classification.} This fully learning-based strategy continuously segments the signal into windows from the outset. Each window is first passed to a binary model that determines whether it corresponds to rest or movement. If classified as rest, no command is sent; if classified as movement, the window is passed to a second model that identifies the specific gesture among the active classes.

\emph{(c) Direct six-class classification.} A single model is trained on six classes (rest and the five active gestures), so activity detection is no longer explicitly separated from gesture classification, and each window is directly classified into one of the six categories.

These three strategies represent three ways of organizing decision-making: a hybrid threshold-based method, a hierarchical two-stage approach, and a unified direct classification. To evaluate them, controlled gesture sequences were recorded and each repetition segmented into successive windows using the previously defined parameters; for each repetition, the percentage of correctly classified windows was computed, and the same recordings were used across strategies for a consistent comparison. Among the three, the threshold-based method was selected based on the experimental results.

\subsubsection{Segmentation, Normalization, and Inference}
\label{subsubsec:inference}

The signal is segmented into \SI{300}{\milli\second} windows (600 samples at \SI{2000}{\hertz}) with a \SI{75}{\milli\second} step, consistent with training (Section~\ref{subsubsec:dataset}). Each window is normalized using the per-channel statistics computed from the training set, so that real-time amplitudes lie in a numerical space comparable to that used during training while preventing information leakage. The normalized window is reshaped into the tensor format expected by the network and passed to the trained CNN, which outputs a predicted class and an associated confidence score. These outputs drive the decision logic and the robotic control module.

\subsection{Simulations and Implementation on the Robotic Arm}
\label{subsec:implementation}

Once a prediction is obtained for a window, it must be translated into a concrete robotic command. This step links the motor intention inferred from the sEMG signals to the actual behavior of the arm, both in simulation and on the real robot. The system uses a simple mapping between hand gestures and Cartesian velocities of the arm, chosen for intuitive and interpretable control:

\begin{tabular}[t]{@{}ll@{\hspace{1em}}ll@{}}
  \textbf{Rest}: & no movement & \textbf{Ulnar deviation}: & motion along $+Y$ \\
  \textbf{Extension}: & motion along $+Z$ & \textbf{Radial deviation}: & motion along $-Y$ \\
  \textbf{Flexion}: & motion along $-Z$ & \textbf{Grip}: & dedicated gripper \\
\end{tabular}

A gesture controls a direction of motion rather than an absolute position: as long as the gesture is recognized, the robot continuously receives a Cartesian velocity consistent with that direction, which suits real-time control based on successive discrete decisions. A sliding "majority vote" approach over the last three windows was implemented, so the command sent to the robot is based on the majority class among the current window and the two preceding ones. This deals with occasional classification errors, reduces artificial changes in direction, and yields more stable control.

The gripper uses a control logic distinct from the Cartesian motion of the arm. Because a single isolated prediction of the grip class could result from a transient error, an additional validation mechanism was introduced. The gripper uses target-position control, with one position for the open state and one for the closed state. When three consecutive detections of the grip gesture occur, the system sends a non-blocking command to move to the new target position. A minimum delay of two seconds between successive state changes is enforced to prevent a single movement from immediately triggering both a closing and a reopening. 

\begin{figure}[htbp]
  \centering
  \begin{subfigure}[t]{0.48\linewidth}
    \centering
    \includegraphics[width=\linewidth]{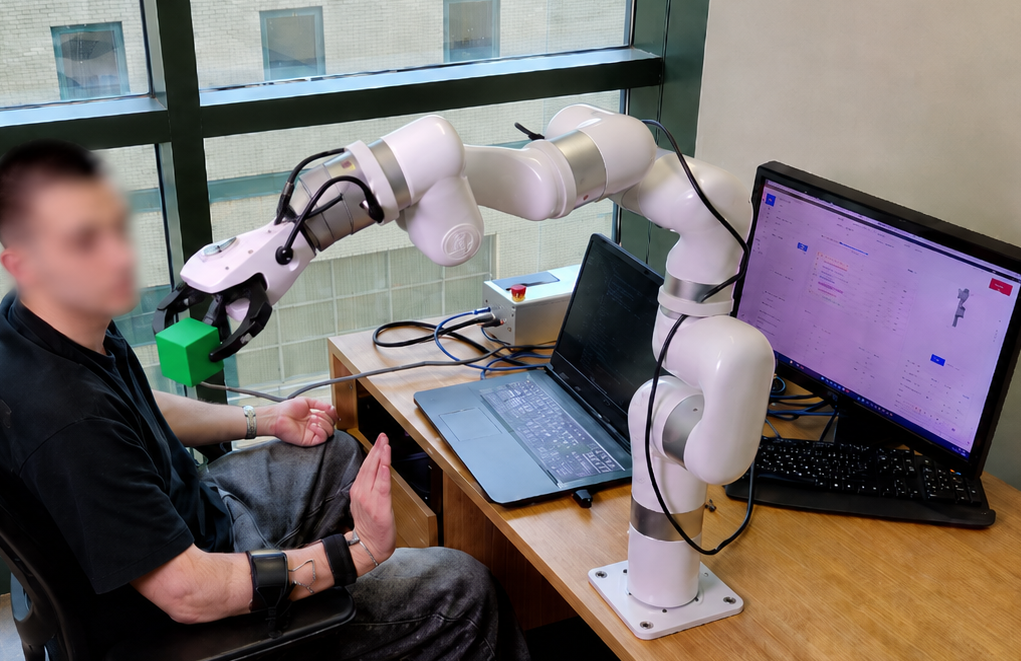}
    \caption{}
    \label{fig:setup_real}
  \end{subfigure}
  \hfill
  \begin{subfigure}[t]{0.49\linewidth}
    \centering
    \includegraphics[width=\linewidth]{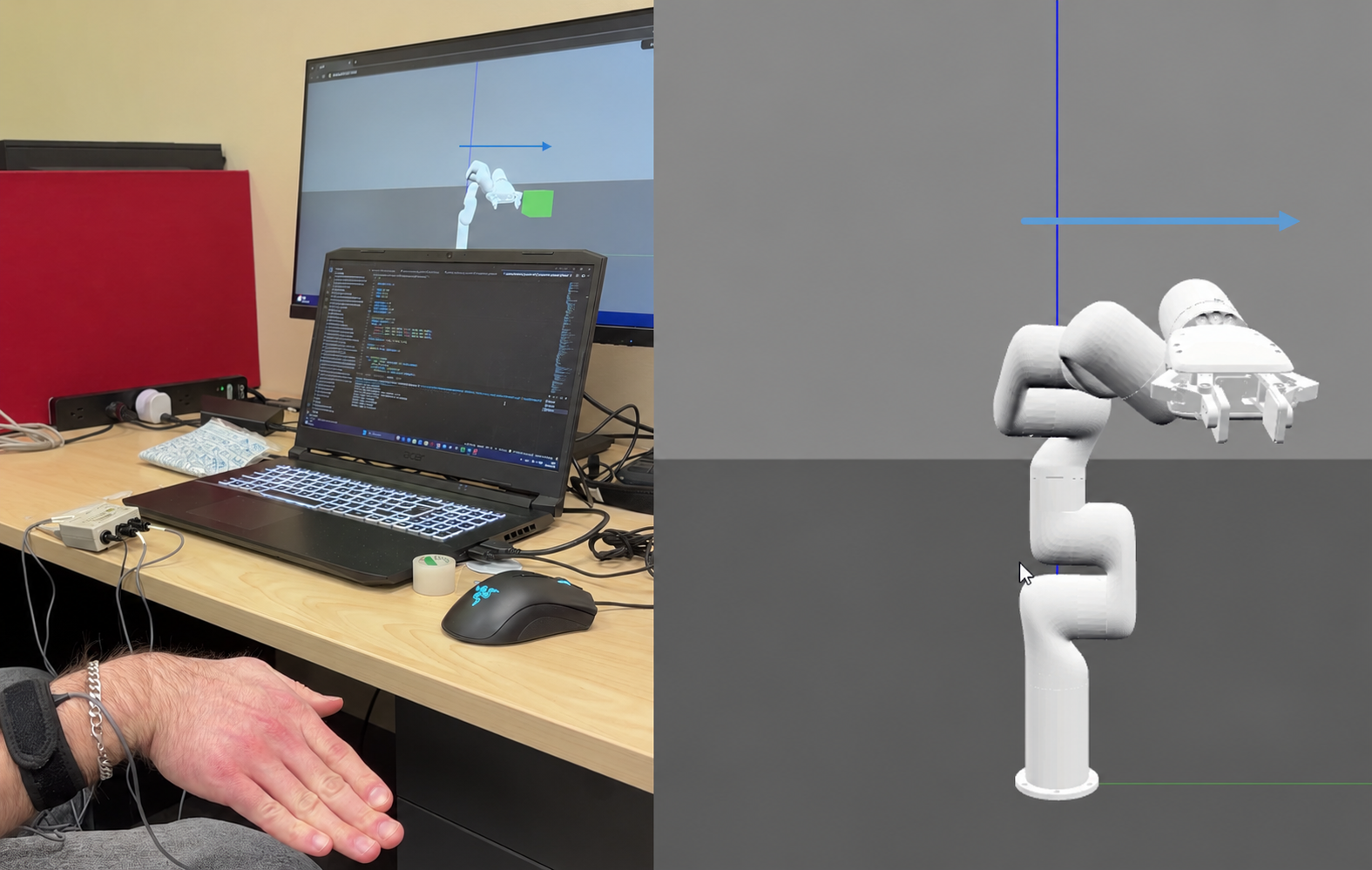}
    \caption{}
    \label{fig:setup_sim}
  \end{subfigure}
  \caption{Experimental setup: (a) the real system, with the surface electrodes on the forearm, the acquisition hardware, and the xArm7 robotic arm; and (b) the simulated xArm7 with the target-zone task in the Swift environment.}
  \label{fig:setup}
\end{figure}

Before integration on the real robot, the commands were tested in a simulated environment based on Robotics Toolbox and Swift, using the kinematic model of the xArm7. The control relies on simple Cartesian control: for a desired Cartesian velocity, the Jacobian is computed at the current joint configuration and the corresponding joint velocities are estimated, applied to the robot model, and integrated over time to update the configuration. The simulation made it possible to validate the gesture-to-command mapping, evaluate the majority-voting mechanism, visually assess control stability, and experiment with gripper strategies without risking damage to the hardware. A target-zone system was also implemented: green cubes were displayed in the workspace, and the objective was to guide the end-effector into these zones; once reached, a zone disappeared and a new one appeared, providing a concrete task for evaluating control performance. After validation in simulation, the system was transferred to a real xArm7 from UFactory. Once initialized in a safe position, the robot is set to Cartesian velocity control mode, and for each new decision produced by the pipeline, a Cartesian velocity corresponding to the majority gesture is sent to the robot. This faithfully reproduces the logic validated in simulation, replacing the internal model dynamics with calls to the real robot SDK, so that the overall system remains conceptually identical with only the final control interface modified. Figure~\ref{fig:setup} shows the two platforms used for validation: the real experimental setup (Figure~\ref{fig:setup_real}) and the simulated xArm7 environment (Figure~\ref{fig:setup_sim}). 

\subsection{Performance Evaluation}
\label{subsec:evaluation}

Evaluation is not limited to classification performance. In robotic teleoperation, it is equally important to assess the latency of the pipeline, the stability of the decisions, and the overall quality of the interaction.

The classification models were evaluated using standard supervised-learning metrics: overall accuracy, precision, recall, F1-score, and the confusion matrix. These quantify overall performance and help identify confusions between specific gestures. Evaluation was performed both on the test set and on sEMG signals acquired in real time and recorded for offline testing, in order to reflect a realistic scenario in which the model must generalize to new measurement conditions.

In a real-time system, latency is a critical criterion, as a correct but late command degrades the interaction. For each window, latency is measured as the time elapsed between the beginning of the window and the moment the corresponding command is sent to the robot. This metric reflects the temporal cost of the pipeline, including window accumulation, preprocessing, inference, decision-making, and command execution.

Beyond quantitative metrics, control quality is evaluated qualitatively through observations during real-time tests, in simulation and on the real robot. Motion stability is assessed to ensure the commands produce consistent movements without excessive oscillations or erratic direction changes, aided by the majority vote over successive predictions. Control smoothness is evaluated by observing the continuity of trajectories during simple gestures, with attention to the absence of jerky motions or abrupt velocity changes. Finally, the overall behavior is evaluated during real-time gesture sequences to assess the consistency between the user's intended actions and the robot's responses, and to identify potential classification errors or control inconsistencies.

\section{Results}
\label{sec:results}

This section first presents the results that guided the methodological choices---onset detection, windowing parameters, classification architecture, and the muscle activation detection strategy---and then the performance of the final real-time pipeline in terms of classification accuracy, latency, and robotic control behavior. Confusion matrices are reported as row-normalized heatmaps in which each cell shows the sample count and, below it, the corresponding percentage of the true class.

\subsection{Methodological Comparisons}
\label{subsec:comparisons}

\subsubsection{Onset Detection Methods}
\label{subsubsec:res_onset}

\begin{figure}[h!]
  \centering
  \includegraphics[width=0.8\linewidth]{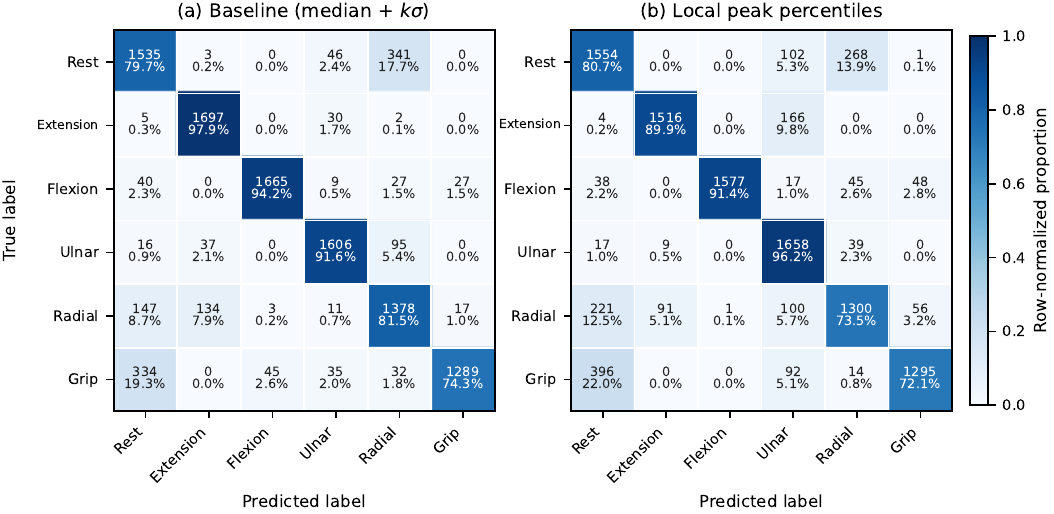}
  \caption{Confusion matrices for onset-detection methods: (a) the baseline method (median $+\,k\sigma$) and (b) the local-peak-percentile}
  \label{fig:onsetcm}
\end{figure}

Two muscle-activation detection approaches were compared: a robust baseline (median $+\,k\sigma$) and a method based on local peak percentiles. The evaluation considered both segmentation behavior (fallbacks) and the impact on the classification model trained on the resulting windows. Regarding segmentation statistics (Table~\ref{tab:fallback}), the robust baseline method exhibits a higher fallback rate (16.0\%, i.e., 160 segments) than the percentile-based method (3.9\%, i.e., 39 segments), indicating that the baseline method is more restrictive and may exclude more portions of the signal. Variability across gestures was also observed, with notably higher fallback rates for some gestures (up to 28.5\% for the grip gesture). However, the impact on classification performance shows the opposite trend (Table~\ref{tab:onset}). The model trained on data segmented with the baseline method achieves an accuracy of 0.8646, compared to 0.8376 for the percentile-based method, and the average F1-scores follow a similar trend. Analysis of the confusion matrices (Figure~\ref{fig:onsetcm}) confirms that the baseline method provides better overall class separation despite greater variability during segmentation, whereas the percentile-based method, though producing fewer fallbacks, yields less discriminative segmentation. These results show that, in this project, stricter segmentation does not degrade performance and may instead improve the quality of the training data. The robust baseline method was therefore selected for the remainder of the pipeline.

\begin{table}[h!]
  \centering
  \small
  \caption{Fallback statistics for the two onset-detection methods (non-rest gestures; 200 trials per gesture, 1000 in total). Onset-time statistics are computed over non-fallback segments.}
  \label{tab:fallback}
  \begin{tabular}{lcc}
    \toprule
     & Baseline (median $+\,k\sigma$) & Local peak percentiles \\
    \midrule
    \multicolumn{3}{l}{\textit{Fallback rate per gesture}} \\
    Extension & 39/200 \; (19.5\%) & 11/200 \; (5.5\%) \\
    Flexion   & 17/200 \; (8.5\%)  & 6/200 \; (3.0\%)  \\
    Ulnar     & 20/200 \; (10.0\%) & 3/200 \; (1.5\%)  \\
    Radial    & 27/200 \; (13.5\%) & 9/200 \; (4.5\%)  \\
    Grip      & 57/200 \; (28.5\%) & 10/200 \; (5.0\%) \\
    \textbf{Total} & \textbf{160/1000 \; (16.0\%)} & \textbf{39/1000 \; (3.9\%)} \\
    \midrule
    \multicolumn{3}{l}{\textit{Onset time (non-fallback)}} \\
    Mean & \SI{526.4}{\milli\second} & \SI{613.5}{\milli\second} \\
    Std  & \SI{184.4}{\milli\second} & \SI{204.9}{\milli\second} \\
    Min  & \SI{100.0}{\milli\second} & \SI{100.0}{\milli\second} \\
    Max  & \SI{977.0}{\milli\second} & \SI{979.5}{\milli\second} \\
    \bottomrule
  \end{tabular}
\end{table}

\begin{table}[h!]
  \centering
  \footnotesize
  \caption{Per-class classification report for the two onset-detection methods (test loss: baseline 0.5178, percentiles 0.6899).}
  \label{tab:onset}
  \begin{tabular}{lcccc}
    \toprule
    Class & Precision & Recall & F1-score & Support \\
    \midrule
    \multicolumn{5}{l}{\textit{Baseline (median $+\,k\sigma$) --- accuracy 0.8646}} \\
    Rest      & 0.7390 & 0.7974 & 0.7671 & 1925 \\
    Extension & 0.9070 & 0.9787 & 0.9415 & 1734 \\
    Flexion   & 0.9720 & 0.9417 & 0.9566 & 1768 \\
    Ulnar     & 0.9246 & 0.9156 & 0.9201 & 1754 \\
    Radial    & 0.7349 & 0.8154 & 0.7731 & 1690 \\
    Grip      & 0.9670 & 0.7429 & 0.8403 & 1735 \\
    Accuracy      &        &        & 0.8646 & 10606 \\
    Macro avg     & 0.8741 & 0.8653 & 0.8664 & 10606 \\
    Weighted avg  & 0.8727 & 0.8646 & 0.8654 & 10606 \\
    \midrule
    \multicolumn{5}{l}{\textit{Local peak percentiles --- accuracy 0.8376}} \\
    Rest      & 0.6969 & 0.8073 & 0.7480 & 1925 \\
    Extension & 0.9381 & 0.8992 & 0.9182 & 1686 \\
    Flexion   & 0.9994 & 0.9142 & 0.9549 & 1725 \\
    Ulnar     & 0.7766 & 0.9623 & 0.8595 & 1723 \\
    Radial    & 0.7803 & 0.7349 & 0.7569 & 1769 \\
    Grip      & 0.9250 & 0.7206 & 0.8101 & 1797 \\
    Accuracy      &        &        & 0.8376 & 10625 \\
    Macro avg     & 0.8527 & 0.8397 & 0.8413 & 10625 \\
    Weighted avg  & 0.8497 & 0.8376 & 0.8387 & 10625 \\
    \bottomrule
  \end{tabular}
\end{table}

\subsubsection{Windowing Parameters}
\label{subsubsec:res_window}

\begin{figure}[h!]
  \centering
  \includegraphics[width=\linewidth]{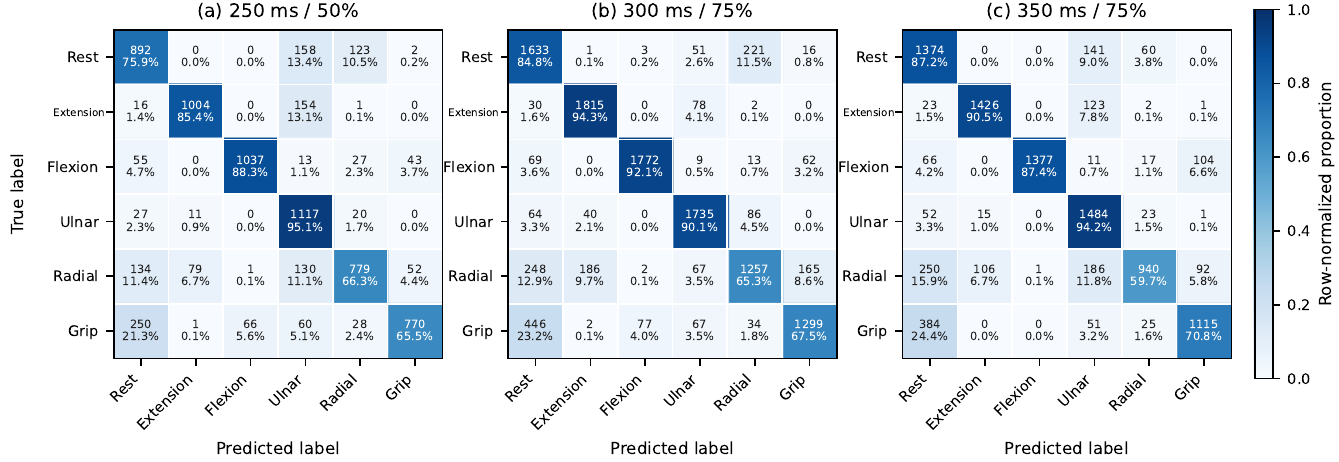}
  \caption{Confusion matrices for the three window sizes: (a) \SI{250}{\milli\second}/50\%, (b) \SI{300}{\milli\second}/75\%, and (c) \SI{350}{\milli\second}/75\%.}
  \label{fig:windowcm}
\end{figure}

The impact of window size and overlap was evaluated by comparing three configurations---\SI{250}{\milli\second}/50\%, \SI{300}{\milli\second}/75\%, and \SI{350}{\milli\second}/75\%---using the same CNN trained on each segmentation (Table~\ref{tab:window}). The \SI{250}{\milli\second}/50\% configuration generates fewer windows (39{,}480 for training) and achieves a test accuracy of 0.7942 with a validation accuracy of 0.9180; it is responsive but less accurate than longer windows. The \SI{300}{\milli\second}/75\% configuration substantially increases the number of windows (64{,}680 for training) and reaches a test accuracy of 0.8235 and a validation accuracy of 0.9397, a notable improvement. The \SI{350}{\milli\second}/75\% configuration achieves the best overall performance, with a test accuracy of 0.8165 and a validation accuracy of 0.9465, benefiting from more temporal information per window. Despite this slightly higher validation performance, the \SI{300}{\milli\second}/75\% configuration was selected for the remainder of the project due to its better test performance and lower latency, reflecting a trade-off between performance and responsiveness. The corresponding confusion matrices are shown in Figure~\ref{fig:windowcm}.

\begin{table}[h!]
  \centering
  \footnotesize
  \caption{Per-class classification report for the three windowing configurations.}
  \label{tab:window}
  \begin{tabular}{lcccc}
    \toprule
    Class & Precision & Recall & F1-score & Support \\
    \midrule
    \multicolumn{5}{l}{\textit{\SI{250}{\milli\second} / 50\% --- accuracy 0.7942}} \\
    Rest      & 0.6492 & 0.7591 & 0.6999 & 1175 \\
    Extension & 0.9169 & 0.8545 & 0.8846 & 1175 \\
    Flexion   & 0.9393 & 0.8826 & 0.9100 & 1175 \\
    Ulnar     & 0.6844 & 0.9506 & 0.7959 & 1175 \\
    Radial    & 0.7965 & 0.6630 & 0.7236 & 1175 \\
    Grip      & 0.8881 & 0.6553 & 0.7542 & 1175 \\
    Macro avg & 0.8124 & 0.7942 & 0.7947 & 7050 \\
    \midrule
    \multicolumn{5}{l}{\textit{\SI{300}{\milli\second} / 75\% --- accuracy 0.8235}} \\
    Rest      & 0.6558 & 0.8483 & 0.7398 & 1925 \\
    Extension & 0.8880 & 0.9429 & 0.9146 & 1925 \\
    Flexion   & 0.9558 & 0.9205 & 0.9378 & 1925 \\
    Ulnar     & 0.8645 & 0.9013 & 0.8825 & 1925 \\
    Radial    & 0.7793 & 0.6530 & 0.7106 & 1925 \\
    Grip      & 0.8424 & 0.6748 & 0.7494 & 1925 \\
    Macro avg & 0.8310 & 0.8235 & 0.8224 & 11550 \\
    \midrule
    \multicolumn{5}{l}{\textit{\SI{350}{\milli\second} / 75\% --- accuracy 0.8165}} \\
    Rest      & 0.6394 & 0.8724 & 0.7379 & 1575 \\
    Extension & 0.9218 & 0.9054 & 0.9135 & 1575 \\
    Flexion   & 0.9993 & 0.8743 & 0.9326 & 1575 \\
    Ulnar     & 0.7435 & 0.9422 & 0.8311 & 1575 \\
    Radial    & 0.8810 & 0.5968 & 0.7116 & 1575 \\
    Grip      & 0.8492 & 0.7079 & 0.7722 & 1575 \\
    Macro avg & 0.8390 & 0.8165 & 0.8165 & 9450 \\
    \bottomrule
  \end{tabular}
\end{table}

\subsubsection{Classification Models}
\label{subsubsec:res_models}

\begin{figure}[h!]
  \centering
  \includegraphics[width=0.8\linewidth]{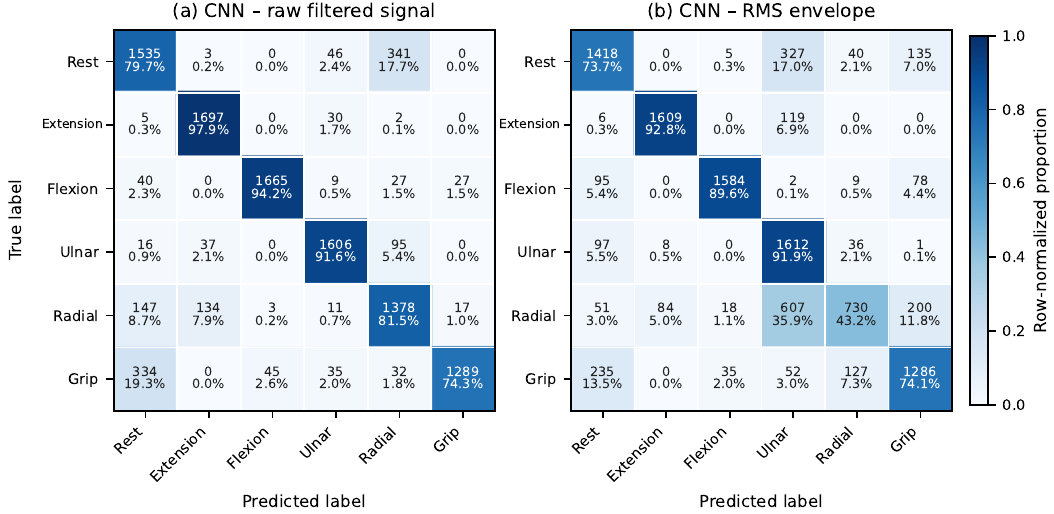}
  \caption{Confusion matrices for the CNN: (a) with the raw filtered signal and (b) with the RMS envelope.}
  \label{fig:cnncm}
\end{figure}

Four configurations were compared on data preprocessed with the selected baseline onset method: a 1D CNN on the raw filtered signal, a CNN on the RMS envelope, and an LSTM on each of these inputs. The CNN on the raw filtered signal achieves the best overall performance, with a test accuracy of 0.8646; class-wise F1-scores are generally high, several gestures exceed 0.90, and the confusion matrix shows good class separation with errors concentrated among specific gesture pairs (Table~\ref{tab:cnn}; confusion matrices in Figure~\ref{fig:cnncm}). The CNN on the RMS envelope yields lower performance, with a test accuracy of 0.7768 and more misclassifications between gestures.

\begin{table}[h!]
  \centering
  \footnotesize
  \caption{Per-class classification report for the CNN with the raw filtered signal versus the RMS envelope (test loss: raw 0.5178, envelope 0.7015).}
  \label{tab:cnn}
  \begin{tabular}{lcccc}
    \toprule
    Class & Precision & Recall & F1-score & Support \\
    \midrule
    \multicolumn{5}{l}{\textit{CNN --- raw filtered signal --- accuracy 0.8646}} \\
    Rest      & 0.7390 & 0.7974 & 0.7671 & 1925 \\
    Extension & 0.9070 & 0.9787 & 0.9415 & 1734 \\
    Flexion   & 0.9720 & 0.9417 & 0.9566 & 1768 \\
    Ulnar     & 0.9246 & 0.9156 & 0.9201 & 1754 \\
    Radial    & 0.7349 & 0.8154 & 0.7731 & 1690 \\
    Grip      & 0.9670 & 0.7429 & 0.8403 & 1735 \\
    Macro avg & 0.8741 & 0.8653 & 0.8664 & 10606 \\
    \midrule
    \multicolumn{5}{l}{\textit{CNN --- RMS envelope --- accuracy 0.7768}} \\
    Rest      & 0.7455 & 0.7366 & 0.7411 & 1925 \\
    Extension & 0.9459 & 0.9279 & 0.9368 & 1734 \\
    Flexion   & 0.9647 & 0.8959 & 0.9290 & 1768 \\
    Ulnar     & 0.5929 & 0.9190 & 0.7208 & 1754 \\
    Radial    & 0.7749 & 0.4320 & 0.5547 & 1690 \\
    Grip      & 0.7565 & 0.7412 & 0.7488 & 1735 \\
    Macro avg & 0.7967 & 0.7754 & 0.7719 & 10606 \\
    \bottomrule
  \end{tabular}
\end{table}

\begin{figure}[h!]
  \centering
  \includegraphics[width=0.8\linewidth]{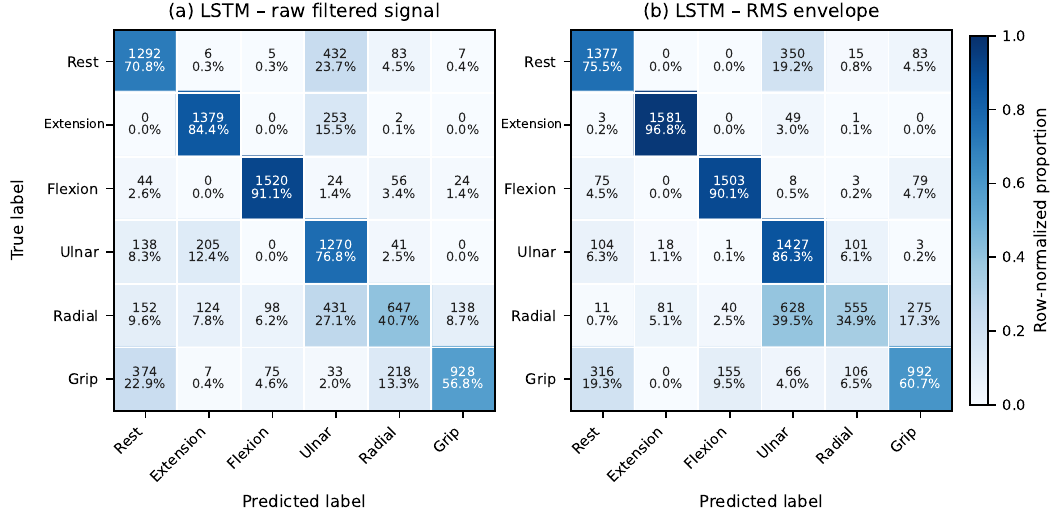}
  \caption{Confusion matrices for the LSTM: (a) with the raw filtered signal and (b) with the RMS envelope.}
  \label{fig:lstmcm}
\end{figure}

Regarding recurrent models, the LSTM on the raw signal reaches a test accuracy of 0.7032 and the LSTM on the RMS envelope 0.7431 (Table~\ref{tab:lstm}; Figure~\ref{fig:lstmcm}); both are significantly lower than the convolutional architectures, and their confusion matrices show greater dispersion with more inter-class errors. Overall, CNN architectures outperform LSTM models here, and the raw filtered signal yields better performance than the RMS envelope. The 1D CNN on the raw signal was therefore selected for the remainder of the project.

\begin{table}[h!]
  \centering
  \footnotesize
  \caption{Per-class classification report for the LSTM with the raw filtered signal versus the RMS envelope (test loss: raw 1.8695, envelope 1.2025).}
  \label{tab:lstm}
  \begin{tabular}{lcccc}
    \toprule
    Class & Precision & Recall & F1-score & Support \\
    \midrule
    \multicolumn{5}{l}{\textit{LSTM --- raw filtered signal --- accuracy 0.7032}} \\
    Rest      & 0.6460 & 0.7079 & 0.6756 & 1825 \\
    Extension & 0.8013 & 0.8439 & 0.8221 & 1634 \\
    Flexion   & 0.8952 & 0.9113 & 0.9031 & 1668 \\
    Ulnar     & 0.5199 & 0.7678 & 0.6200 & 1654 \\
    Radial    & 0.6180 & 0.4069 & 0.4907 & 1590 \\
    Grip      & 0.8459 & 0.5676 & 0.6794 & 1635 \\
    Macro avg & 0.7210 & 0.7009 & 0.6985 & 10006 \\
    \midrule
    \multicolumn{5}{l}{\textit{LSTM --- RMS envelope --- accuracy 0.7431}} \\
    Rest      & 0.7301 & 0.7545 & 0.7421 & 1825 \\
    Extension & 0.9411 & 0.9676 & 0.9541 & 1634 \\
    Flexion   & 0.8846 & 0.9011 & 0.8928 & 1668 \\
    Ulnar     & 0.5645 & 0.8628 & 0.6824 & 1654 \\
    Radial    & 0.7106 & 0.3491 & 0.4682 & 1590 \\
    Grip      & 0.6927 & 0.6067 & 0.6469 & 1635 \\
    Macro avg & 0.7539 & 0.7403 & 0.7311 & 10006 \\
    \bottomrule
  \end{tabular}
\end{table}

\subsubsection{Real-Time Muscle Activation Detection Strategies}
\label{subsubsec:res_strategies}

The three real-time strategies were evaluated on a common set of experimental recordings: (1) threshold-based detection followed by a 5-gesture classifier, (2) two-stage classification (rest vs move followed by 5-gesture classification), and (3) single 6-class classification. For each recorded gesture, the signal was segmented into successive windows and the percentage of correctly classified windows was computed, with the reported results corresponding to the average recognition rates over multiple repetitions (Tables~\ref{tab:threshold}--\ref{tab:sixclass}).

\begin{table}[h!]
  \centering
  \small
  \caption{Threshold-based method (5-gesture classifier): per-gesture precision (\%).}
  \label{tab:threshold}
  \begin{tabular}{lccccc}
    \toprule
     & Extension & Flexion & Ulnar & Radial & Grip \\
    \midrule
    Precision (\%) & 98.16 & 99.86 & 99.61 & 70.98 & 99.33 \\
    \bottomrule
  \end{tabular}
\end{table}

\begin{table}[htbp]
  \centering
  \small
  \caption{Two-stage method: for each gesture, the percentage of windows correctly routed by the rest/move stage and, among the windows detected as ``move'', the percentage of correctly recognized gestures.}
  \label{tab:twostage}
  \begin{tabular}{lcc}
    \toprule
    Class & Detected as move (\%)\textsuperscript{a} & Correctly recognized gesture (\%) \\
    \midrule
    Rest      & 100.00\textsuperscript{b} & --- \\
    Extension & 99.09 & 99.05 \\
    Flexion   & 98.40 & 100.00 \\
    Ulnar     & 99.60 & 99.61 \\
    Radial    & 92.70 & 70.54 \\
    Grip      & 97.98 & 99.73 \\
    \bottomrule
  \end{tabular}

  \vspace{2pt}
  {\footnotesize \textsuperscript{a}For the rest class, this column reports the percentage of windows correctly detected as ``rest''. \textsuperscript{b}Rest detection rate.}
\end{table}

\begin{table}[h!]
  \centering
  \small
  \caption{Direct 6-class method: per-class precision (\%).}
  \label{tab:sixclass}
  \begin{tabular}{lcccccc}
    \toprule
     & Rest & Extension & Flexion & Ulnar & Radial & Grip \\
    \midrule
    Precision (\%) & 99.84 & 98.30 & 99.60 & 100.00 & 67.73 & 97.59 \\
    \bottomrule
  \end{tabular}
\end{table}

All three methods enable reliable recognition, with high rates for most classes. The threshold-based method (5-gesture classification) achieves above 98\% for most gestures, except the radial gesture (70.98\%). The two-stage strategy shows comparable overall performance, with recognition rates for gestures detected as ``move'' ranging from 92\% to 100\% depending on the class, and robust rest detection (100\%). The direct 6-class approach shows similar performance for most classes, with rates above 97\% for several gestures but a noticeable degradation for the radial gesture (67.73\%), confirming a recurring difficulty in classifying this movement. The most frequent errors involve confusions between similar gestures, particularly ulnar and radial movements, as well as certain transitions involving the grip gesture.

In addition, qualitative observations were conducted during simulation-based tests. All three methods enable generally functional robot control, with smooth movements in most cases, and the average latency is similar across approaches, ranging from approximately \SI{314}{} to \SI{325}{\milli\second}, mainly reflecting the chosen window size and the majority-voting mechanism. However, the threshold-based method provides stable control with few visible errors, whereas the two-stage approach may introduce undesired gripper activations at the end of movements, and the 6-class method exhibits more classification errors and less smooth motion in certain cases. For these reasons, the threshold-based detection method followed by a 5-gesture classifier was selected.

\subsection{Final Selected System}
\label{subsec:finalsystem}

\subsubsection{Performance of the Final Model}
\label{subsubsec:res_final}

\begin{figure}[h!]
  \centering
  \includegraphics[width=0.5\linewidth]{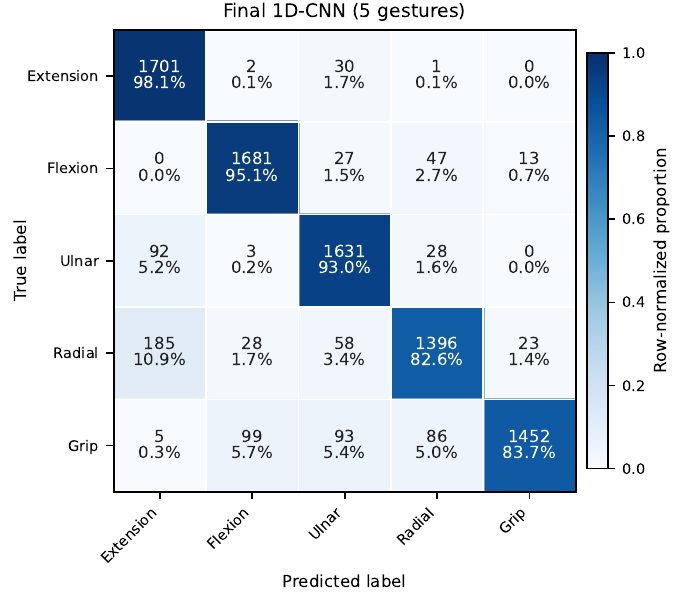}
  \caption{Confusion matrix of the final selected 1D-CNN (5-gesture) model on the test set.}
  \label{fig:finalcm}
\end{figure}

The model selected for the real-time system is a 1D CNN trained on the raw filtered sEMG signal for 5-gesture classification, evaluated both on a test set derived from the dataset and on independently recorded experimental data. On the test set, the model achieves an overall accuracy of 0.9055 with a loss of 0.3738; class-wise F1-scores range from 0.8596 to 0.9388 (Table~\ref{tab:final}). Extension and flexion exhibit the best performance, while some confusions appear for lateral gestures (ulnar and radial), as shown in the confusion matrix (Figure~\ref{fig:finalcm}).

\begin{table}[htbp]
  \centering
  \footnotesize
  \caption{Per-class classification report for the final 1D-CNN (5-gesture) model on the test set (test loss 0.3738).}
  \label{tab:final}
  \begin{tabular}{lcccc}
    \toprule
    Class & Precision & Recall & F1-score & Support \\
    \midrule
    Extension & 0.8578 & 0.9810 & 0.9153 & 1734 \\
    Flexion   & 0.9272 & 0.9508 & 0.9388 & 1768 \\
    Ulnar     & 0.8869 & 0.9299 & 0.9079 & 1754 \\
    Radial    & 0.8960 & 0.8260 & 0.8596 & 1690 \\
    Grip      & 0.9758 & 0.8369 & 0.9010 & 1735 \\
    \midrule
    Accuracy      &        &        & 0.9055 & 8681 \\
    Macro avg     & 0.9087 & 0.9049 & 0.9045 & 8681 \\
    Weighted avg  & 0.9088 & 0.9055 & 0.9049 & 8681 \\
    \bottomrule
  \end{tabular}
\end{table}

Under more realistic conditions, additional tests were conducted on experimentally acquired sEMG signals; for each gesture repetition, the percentage of correctly classified windows was computed (Table~\ref{tab:expdata}). The average results are 98.16\% for extension, 99.86\% for flexion, 99.61\% for the ulnar gesture, and 99.33\% for grip. The radial gesture again shows lower performance, with an average of 70.98\%, confirming the difficulty observed in earlier stages. Error analysis indicates that the most frequent confusions occur between lateral gestures, particularly ulnar and radial, and during certain transitions involving the grip gesture. Overall, the final model achieves very strong classification performance with high robustness across most gestures, including under experimental conditions.

\begin{table}[h!]
  \centering
  \small
  \caption{Percentage of correctly classified windows for each gesture, measured over ten real-time test repetitions on experimentally acquired sEMG data.}
  \label{tab:expdata}
  \begin{tabular}{lccccc}
    \toprule
    Test \# & 1-Extension & 2-Flexion & 3-Ulnar & 4-Radial & 5-Grip \\
    \midrule
    1  & 96.1 & 100  & 100  & 88.0 & 100  \\
    2  & 100  & 100  & 100  & 97.2 & 100  \\
    3  & 100  & 100  & 100  & 100  & 100  \\
    4  & 97.3 & 100  & 100  & 72.2 & 100  \\
    5  & 100  & 98.6 & 100  & 95.8 & 97.3 \\
    6  & 94.8 & 100  & 96.1 & 37.8 & 100  \\
    7  & 100  & 100  & 100  & 63.5 & 98.7 \\
    8  & 100  & 100  & 100  & 46.6 & 97.3 \\
    9  & 98.6 & 100  & 100  & 21.9 & 100  \\
    10 & 94.8 & 100  & 100  & 86.8 & 100  \\
    \midrule
    \textbf{Mean} & \textbf{98.16} & \textbf{99.86} & \textbf{99.61} & \textbf{70.98} & \textbf{99.33} \\
    \midrule
    Most common mistake & Class 3--4 & & & Class 1--5 & \\
    \bottomrule
  \end{tabular}

  \vspace{2pt}
  {\footnotesize All values are percentages (\%). ``Most common mistake'' gives the most frequently confused gesture pair, using the class indices in the header (3--4: ulnar/radial; 1--5: extension/grip).}
\end{table}

\subsubsection{System Latency Analysis}
\label{subsubsec:res_latency}

\begin{figure}[h!]
  \centering
  \includegraphics[width=0.75\linewidth]{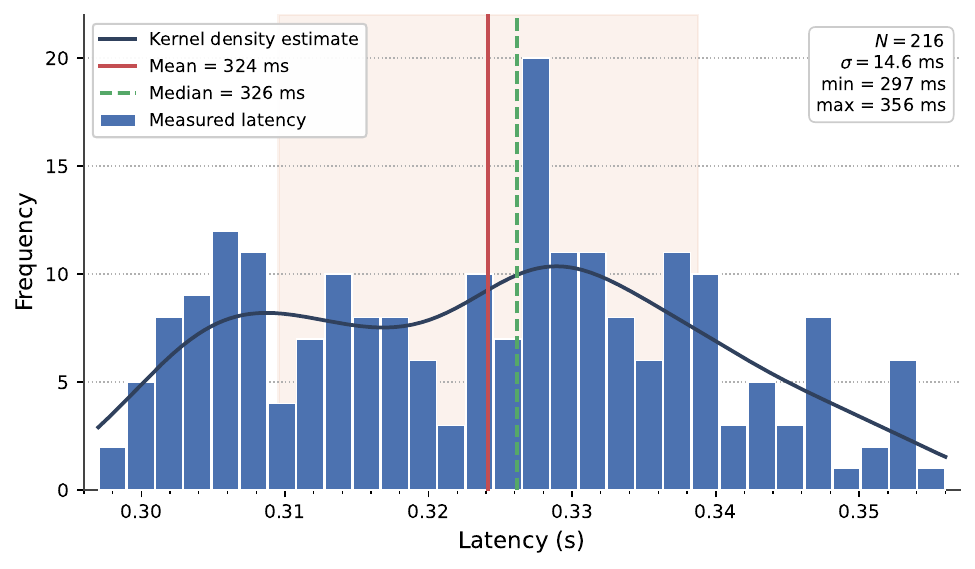}
  \caption{Histogram of the system latency distribution measured during real-time robot control.}
  \label{fig:latency}
\end{figure}

System latency was evaluated during several real-time testing sessions with the robot, defined as the interval between the beginning of the signal window used for prediction and the transmission of the corresponding command. The results show an average latency of \SI{0.324}{\second} with a standard deviation of \SI{0.0146}{\second}, indicating relatively low variability (Figure~\ref{fig:latency}). The extreme values range from a minimum of \SI{0.297}{\second} to a maximum of \SI{0.356}{\second}, with a median of \SI{0.326}{\second}. The distribution exhibits limited dispersion around the mean, reflecting stable behavior. This latency is primarily determined by the \SI{300}{\milli\second} window size and the majority-voting decision mechanism, and it remains compatible with real-time operation while maintaining a balance between responsiveness and classification robustness.

\subsubsection{System Behavior Under Real Conditions}
\label{subsubsec:res_realconditions}

The system was evaluated both in simulation and on the real robot during real-time testing. Controlling the robot from sEMG signals proved functional and stable, with movements consistent with the executed gestures. The generated motions are generally smooth and continuous, particularly for the main gestures (extension, flexion, ulnar), and the system reliably detects and interprets movement intentions, although a slight delay can be observed at the onset of activation due to system latency. The grip action is correctly handled through the robust consecutive-vote detection logic, which helps prevent unintended triggering, although occasional unintended activations may still occur, particularly at the end of movements. Some classification errors persist, mainly for lateral gestures (especially radial), which may cause occasional incorrect movements; however, these errors remain limited and do not prevent overall control. Overall, the system enables intuitive and responsive robot control, with similar behavior in simulation and on the real robot.
\section{Discussion}
\label{sec:discussion}

The results demonstrate that sEMG-based robotic teleoperation is feasible in a real-time setting, provided that the entire pipeline from signal acquisition to robotic control is carefully structured. Beyond raw classification performance, this project highlights the importance of balancing accuracy, robustness, implementation simplicity, and system responsiveness. The following discussion interprets the results, contextualizes the methodological choices, and outlines the main limitations and future perspectives.

\subsection{Interpretation of the Results and Validation of Methodological Choices}
\label{subsec:disc_interpretation}

The methodological comparisons show that the selected choices rest on consistent experimental results rather than theoretical considerations alone. For onset detection, the robust baseline method (median and MAD) led to better classification performance than the percentile-based peak method (test accuracy 0.8646 versus 0.8376; Table~\ref{tab:onset}), despite a markedly higher fallback rate (16.0\% versus 3.9\%; Table~\ref{tab:fallback}). This may appear counterintuitive, as a more permissive method could seem advantageous by retaining more segments; however, stricter segmentation appears to have better isolated the relevant portions of the signal, providing more consistent training examples. The confusion matrices in Figure~\ref{fig:onsetcm} support this interpretation, showing sharper diagonal dominance and therefore cleaner class separation for the baseline method. This suggests that, in sEMG classification, the quality of segmentation matters more than the mere quantity of retained data.

A similar trade-off appears in the windowing analysis (Table~\ref{tab:window}). Window size strongly affects classifier performance: the \SI{250}{\milli\second}/50\% configuration is the most responsive but the least accurate (test accuracy 0.7942), whereas longer windows improve stability at the cost of latency. The \SI{300}{\milli\second}/75\% window achieves the best test accuracy (0.8235) while keeping the segmentation delay low, and it was therefore retained even though the \SI{350}{\milli\second}/75\% configuration reached a marginally higher validation accuracy but a lower test accuracy (0.8165). This choice is particularly relevant here, since the goal is not only to optimize offline accuracy but also to ensure sufficiently smooth behavior in a real-time control system.

The comparison of models (Tables~\ref{tab:cnn} and~\ref{tab:lstm}; Figures~\ref{fig:cnncm} and~\ref{fig:lstmcm}) confirms the relevance of a 1D CNN applied to the raw filtered signal, which outperformed both the RMS-envelope input (0.8646 versus 0.7768) and the two LSTM variants (0.7032 and 0.7431). Notably, replacing the raw signal with its envelope degrades the lateral gestures most severely (the radial recall drops from 0.8154 to 0.4320 (Table~\ref{tab:cnn})) which indicates that the information required to discriminate these gestures is not limited to a global activity envelope but is embedded in the fine temporal structure of the signal. 1D convolutions extract this information effectively without the added complexity, and higher minimum latency, of a recurrent model. Finally, the comparison of the three real-time strategies (Tables~\ref{tab:threshold}--\ref{tab:sixclass}) shows that no single approach is universally optimal, but some are better suited to the intended application. The threshold-based strategy provides a clear separation between activity detection and motion recognition, contributing to its robustness; the two-stage approach offers a fully learning-based architecture at the cost of increased complexity; and the direct six-class classification simplifies the pipeline but requires the model to handle both rest detection and gesture recognition simultaneously. All three reach high recognition rates for most gestures, yet all degrade specifically on the radial gesture (70.98\%, 70.54\%, and 67.73\% respectively), and they do not provide the same stability under control conditions. The best method is therefore not necessarily the one with the highest offline accuracy, but the one that maintains reliable functional behavior within the complete system.

\subsection{Analysis of Real-Time System Behavior}
\label{subsec:disc_realtime}

A major component of this work is the full integration of the pipeline within a real-time control loop, first in simulation and then on the real robot. System performance can no longer be assessed solely through quantitative metrics; it must also be evaluated in terms of control quality, motion consistency, and perceived responsiveness. The measured latency (Figure~\ref{fig:latency}), averaging \SI{0.324}{\second} with a standard deviation of \SI{0.0146}{\second} and ranging from \SIrange{0.297}{0.356}{\second}, is consistent with the design: a significant portion originates from the \SI{300}{\milli\second} window size, which imposes a minimum delay before a decision can be made. Nevertheless, this delay remains low enough for practical real-time control, and its limited dispersion indicates stable, predictable timing. The satisfactory behavior of the robot in both simulation and real-world experiments confirms that the model's predictions can be effectively translated into coherent commands. The observed stability is largely due to the majority vote over the last three windows, which smooths occasional classification errors and limits rapid command fluctuations, at the expense of slightly reduced responsiveness to rapid changes---a trade-off appropriate here, where stable motion is preferable to reactive but unstable control. Regarding the gripper, activation based on multiple consecutive detections combined with a minimum delay between actions effectively prevents unintended triggering while remaining compatible with real-time control, demonstrating that the pipeline can incorporate discrete binary actions without compromising overall performance.

\subsection{System Limitations}
\label{subsec:disc_limitations}

Several limitations must be highlighted. The first is the difficulty in distinguishing certain similar gestures, particularly lateral movements such as ulnar and radial deviations. This is a recurring issue across all confusion matrices and is confirmed by the final model, whose radial gesture has both the lowest recall (0.8260; Table~\ref{tab:final}) and, in the confusion matrix (Figure~\ref{fig:finalcm}), the largest off-diagonal mass, concentrated on the neighboring lateral and grip classes. The effect is even more pronounced under real-time conditions: the per-repetition experimental accuracy of the radial gesture averages only 70.98\% and drops as low as 21.9\% in individual trials (Table~\ref{tab:expdata}), by far the weakest of all gestures. This behavior may be explained by the physiological similarity of the muscle activations associated with these movements, the limited number of channels, and the sensitivity of the system to precise electrode placement. A second limitation is the dependence on experimental conditions: strong consistency between the training dataset and real-time acquisitions is essential, so electrode placement, contact quality, noise levels, and the user's muscle condition can significantly affect performance. The large repetition-to-repetition variability of the radial gesture in Table~\ref{tab:expdata} illustrates this sensitivity and represents a key challenge for broader deployment. Latency, although reasonable, also remains a constraint; even stable latency may be perceived by the user, especially during rapid gesture transitions, and reducing it would require revisiting the trade-off between window size, decision frequency, and classification robustness. Finally, the evaluation remains partial: while the project demonstrates feasibility, it does not yet constitute a comprehensive validation under diverse operating conditions, so the results are not sufficient to conclude full robustness across all scenarios.

\subsection{Future Work}
\label{subsec:disc_future}

Several avenues emerge from this study. Experimentally, extending the evaluation to multiple users would allow a better assessment of robustness with respect to inter-subject variability. Another important direction is the evaluation of usability: aspects such as the ease of producing reliable commands, perceived responsiveness, control naturalness, and fatigue from repeated use were not formally investigated, yet they are essential from an application perspective and could be analyzed through dedicated protocols or user questionnaires. From an algorithmic standpoint, alternative strategies or more advanced models capable of better handling sEMG variability could be explored, and improved handling of gesture transitions and uncertain states could enhance the smoothness of control. A particularly promising avenue lies in hybrid control approaches combining sEMG with other sources of information e.g., integrating vision sensors or environmental perception to associate the user's intention with an understanding of the context, thereby improving accuracy and robustness and facilitating the execution of more complex tasks.

\section{Conclusion}
\label{subsec:conclusion}

Upper limb motor impairments represent a major issue for autonomy and quality of life, particularly for individuals with long-term functional deficits, and the development of intuitive, reliable control interfaces is key to making assistive robotic devices truly usable. This work addressed this problem by proposing a human--machine interface based on sEMG signals that enables telecontrol of a robotic arm through simple hand and wrist gestures.

One of the main contributions is the implementation of a complete real-time processing pipeline, from sEMG acquisition to effective robot control. Unlike many studies focused only on offline classification, this project integrated all the steps required for operational functioning including signal filtering, muscle activation detection, window segmentation, normalization, convolutional neural network inference, and generation of robotic commands, and validated them both in simulation and on a real robot. The results confirm the relevance of the methodological choices: the 1D CNN applied to the raw filtered signal proved the most effective among the tested architectures, achieving high accuracy on the test data and excellent performance on independently acquired experimental data. The analysis of the real-time strategies also showed that several approaches are viable, each with advantages and trade-offs in robustness and simplicity, and the system latency of around \SI{0.32}{\second} is consistent with the selected segmentation parameters and compatible with real-time use for simple tasks.

Beyond numerical performance, the tests validated the overall behavior of the system in a robotic telecontrol context: the robot can be controlled coherently from the user's gestures, with globally smooth and stable movements, and the majority vote over successive windows proved particularly effective in limiting occasional errors. The consistency between simulation and real-robot results strengthens the robustness of the pipeline and demonstrates the feasibility of such a system in an applied setting. Nevertheless, the classification of certain gestures, particularly lateral movements, remains difficult because of the similarity of the associated muscle activations; the system remains sensitive to acquisition conditions; latency, although stable, is inherent to temporal segmentation; and the evaluation remains partial. These limitations open several perspectives, from more in-depth evaluation under varied conditions and with more users to improvements in the classification models, online adaptation, rest-state detection, and hybrid control approaches combining sEMG with complementary sensing. Overall, this work demonstrates that an sEMG-based robotic telecontrol system can operate in real time with satisfactory performance, providing a solid basis for more advanced assistive systems that combine machine learning, signal processing, and robotics to improve the interaction between humans and assistive technological devices.

\bibliographystyle{unsrtnat}
\bibliography{refs}

\end{document}